\documentclass[10pt,twocolumn,a4paper]{article}

\usepackage[left=15mm, right=15mm, top=15mm, bottom=15mm]{geometry}

\usepackage{flushend}
\usepackage{indentfirst}
\usepackage{graphics}
\usepackage{amsmath}
\usepackage{graphicx}
\usepackage{epstopdf}
\usepackage{float}

\usepackage{url}
\usepackage{amsmath,amssymb}
\usepackage[linesnumbered,ruled,vlined]{algorithm2e}

\mathchardef\mhyphen="2D 

\usepackage{pgf,tikz}
\usepackage{mathrsfs}
\usetikzlibrary{arrows}

\usepackage[font=footnotesize,labelfont=bf]{caption}
\usepackage[labelsep=period]{caption}

\usepackage{subcaption}

\begin{document}

\title{\huge \textbf{Detecting Point Outliers Using Prune-based Outlier Factor (PLOF)}}

\twocolumn[
\begin{@twocolumnfalse}

\author{\textbf{Kasra Babaei}$^{1}$, \textbf{ZhiYuan Chen}$^{1,*}$, \textbf{Tomas Maul}$^{1}$\\\\
\footnotesize $^{1}${School of Computer Science, University of Nottingham Malaysia,}\\
\footnotesize $^{*}$Corresponding Author: Zhiyuan.Chen@nottingham.edu.my}

\date{}

\maketitle


\end{@twocolumnfalse}
]

\noindent \textbf{\large{Abstract}} \hspace{2pt} Outlier detection (also known as anomaly detection or deviation detection) is a process of detecting data points in which their patterns deviate significantly from others. It is common to have outliers in industry applications, which could be generated by different causes such as human error, fraudulent activities, or system failure.  Recently, density-based methods have shown promising results, particularly among which Local Outlier Factor (LOF) is arguably dominating. However, one of the major drawbacks of LOF is that it is computationally expensive. Motivated by the mentioned problem, this research presents a novel pruning-based procedure in which the execution time of LOF is reduced while the performance is maintained. A novel Prune-based Local Outlier Factor (PLOF) approach is proposed, in which prior to employing LOF, outlierness of each data instance is measured. Next, based on a threshold, data instances that require further investigation are separated and LOF score is only computed for these points. Extensive experiments have been conducted and results are promising. Comparison experiments with the original LOF and two state-of-the-art variants of LOF have shown that PLOF produces higher accuracy and precision while reducing execution time.\\

\noindent \textbf{\large{Keywords}} \hspace{2pt} Anomaly Detection, Deviation Detection, Local Outlier Factor, Outlier Detection,\\

\noindent\hrulefill

\section{\Large{Introduction}}
\par Outlier detection (also known as anomaly detection or deviation detection) is a process of detecting data points in which their patterns deviate significantly from others. This definition may vary slightly based on the area that it is used \cite{Chandola:2009:ADS:1541880.1541882}. Outlier detection has been employed in various application domains such as financial, telecommunications, medical, and commercial industries \cite{Ha201415}. 

\par According to \cite{Chandola:2009:ADS:1541880.1541882}, outliers are categorised into three major types, namely point outliers, collective outliers, and contextual outliers. The methods used for detecting these three types are very different from each other and often it is impossible to employ the same method on the other two. In this paper, we have proposed an approach for detecting point outliers and detecting contextual or collective outliers is out of the scope of this paper.

\par Previous studies have proposed several methods for outlier detection, and it is hard to find a universal categorisation for all the methods. However, \cite{Tang2017171} categorise them into the following four groups: distribution-based, distance-based, clustering-based, and density-based methods. While in distribution-based methods the assumption is that outliers are generated from a distribution that deviates from the others, in distance-based methods the assumption is that outlier points tend to have a distance above a threshold from a certain proportion of the others. In clustering-based methods, the data instances are clustered and instances that do not belong to any cluster are considered as the outlier. Lastly, a density-based method tries to compute the density of each data instance and then compares the density with the density of the neighbourhood. Instances that the ratio of their density and the density of their neighbours are above a threshold are the outlier. 

\par \cite{Breunig:2000:LID:335191.335388} proposed a density-based method called Local Outlier Factor (LOF) that generates a value for each data instance that tells its outlierness. Instead of using a threshold, top-$n$ data instances that have the highest LOF value are considered as outliers. LOF has shown promising results, however, one of the major drawbacks of LOF is that it is computationally very expensive, i.e., $O(N^2)$ \cite{8291963}. Therefore, researchers have implemented pruning-based methods to reduce $N$, i.e., decrease computation cost. However, it has been realised that it will be achieved only at the cost of debilitating performance of outlier detection.

\par In this paper, we propose an outlier detection method based on LOF that would decrease the execution time by bringing down the complexity. Instead of computing LOF value for all the data instances, we perpetuate a preliminary step in which density of each data instance is estimated and then pruning is carried out based on the median of the density of data instances. Finally, LOF is only computed for the remaining data instances. The second contribution is maintaining the performance. As mentioned in Previous Studies, several pruning-based approaches are proposed in the literature; however, most of these works reduce execution time by sacrificing the performance. In summary, the contributions of this paper are two-fold:
\begin{itemize}
\label{item:contribution}
  \item Introducing a new density estimation that reduces the execution time of LOF.
  \item Maintaining the performance of outlier detection.
\end{itemize}


\section{\Large{Related Studies}}
\label{sec:previous}
\par \cite{Breunig:2000:LID:335191.335388} proposed a density-based method known as LOF in which a value is given to each data instance that shows the outlierness for that individual. Then, the top-$n$ points with the highest LOF value are considered as outliers. Often, the points with LOF value above $1$ are regarded as outlier \cite{6460620}. A major disadvantage of LOF is its complexity, i.e., $O(N^2)$ where $N$ is the size of the data \cite{Sinha2018}. The authors of \cite{1214939} proposed three improvements: $LOF'$, $LOF''$ and $GridLOF$ to simplify the computation of the original LOF. In $LOF'$, merely the ratio of $MinPts{-}dist$ of the query object and its neighbours are used to compute the LOF value. They argued that this ratio is sufficient and it is needless to compute the reachability distance and local reachability density. And, in $LOF''$, they employed two $MinPts$ instead of just one to enhance the performance of LOF. Their last enhancement, $GridLOF$, is a pruning-based approach that removes dense areas so that computing LOF for data instances inside the dense areas is not required anymore. However, the drawback of $GridLOF$ is that it requires manual grid setting, which is not always feasible. 
\par In previous work, \cite{6460620}, the authors randomly divided the dataset into chunks and then determined the nearest neighbours of each data instance only based on the instances within the chunk that the query point belongs to. Next, using the computed nearest neighbours they generated LRD and LOF for all the data instances. However, this requires a precise neighbour selection and also not all the outliers may be detected. 
\par Another pruning approach was proposed by \cite{Pamula2012210} in which a clustering algorithm was applied in advance, and then dense clusters are pruned based on this assumption that they contain no outliers. Next, they employed a Local Distance-based Outlier Factor (LDOF) to the sparse and small clusters to detect outliers. Earlier, \cite{5734938} proposed a similar method that after clustering the dataset by using $k$-means, instances that are close to the centroid of the cluster, which they belong to, were pruned, and LDOF was used merely on instances that were away from the centroid, i.e., outside of a predefined radius. 
\par \cite{Rizk2015} claimed that their approach can not only reduce the calculation rate of LOF but also minimises the false negative rate. Their approach has two stages. In the first stage, the data instances are clustered using $k$-medoids, which they believe its robustness against outliers and noise is more than $k$-means. Next, after computing a local cut-off value based on the size of the cluster, instances that are outside of the radius are considered as potential outliers. In the final stage, LOF only computed for potential outliers that were obtained from the previous stage. 
\par The authors of \cite{Poddar2018} presented a generic method for reducing the execution time of many density-based and distance-based outlier detection algorithms. In their method, a new density estimation called $devToMean$ was introduced that based on its value normal data instances were pruned, then the outlier detection method was applied only on the rest of the data. However, the computation of $devToMean$ is expensive, therefore, they used $k$-means to divide the dataset into small clusters, and then calculated the value of $devToMean$ for each instance within its cluster.


\section{\Large{Proposed Method}}
\label{sec:proposed_method}
\subsection{\normalsize \textbf{Local Outlier Factor}}
\label{sec:LOF}
\par This algorithm was first introduced by \cite{Breunig:2000:LID:335191.335388}. Local outlier factor (LOF) is a density-based algorithm that gives a score to each data point based on the density of its local neighbourhood. This score represents the degree by which the data point deviates from its local neighbourhood.

\par For any positive integer $k$, let $k$-distance be the distance between object $p$ and its $k$ nearest objects. The $k$-distance-neighbourhood of object $p$ contains all the objects whereby their distance from $p$ are not greater than the $k$-distance, i.e. $N_{k{\mhyphen}dist(p)}=\{p'{\mid}p'{\in}D,dist(p,p')\leq{dist_{k}(p)}\}$. 

\par The $k$-distance and $k$-distance-neighbourhood are used to compute the \textit{reachability distance} of an object. The reachability distance of object $p$ with respect to object $o$ is defined as:

\begin{equation}
	reach{\mhyphen}dist_{k}(p,o)=max\{k{\mhyphen}distance(o),d(p,o)\}
	\label{equ:reach-dist}
\end{equation}

\par The authors of \cite{Breunig:2000:LID:335191.335388} proposed a specific instantiation of $k$ for outlier detection, which is derived from density-based algorithms. Density-based algorithms often have two parameters. The first parameter, known as $MinPts$, determines the minimum number of points that are needed. There is also the second parameter that defines a volume. The two parameters are used to define a density threshold for algorithms. However, LOF only utilises $MinPts$ and dynamically specifies the density by employing $reach{\mhyphen}dist_{MinPts}(p,o)$, for $o{\in}N_{MinPts}(p)$. By specifying $MinPts$, and given the $reach-dist$, the local reachability density of an object is defined as:

\begin{equation}
	{\scriptstyle LRD_{MinPts}(p)=1/}\Bigg(\frac{\mathlarger{\sum}_{\scriptscriptstyle{o\in_{N_{MinPts}(p)}}}{\scriptscriptstyle{reach{\mhyphen}dist_{MinPts}(p, o)}}}{\scriptscriptstyle{|N_{MinPts}(p)|}}\Bigg)
	\label{equ:lrd}
\end{equation}

which is the inverse of the average reachability distance based on the parameter $MinPts$ of object $p$. Local reachability can be $\infty$ in the case that every reachability distance in the summation is 0, which may happen when there are data points that have the same spatial location of object $p$, i.e. duplicates of $p$.

\par Local outlier factor, which represents the deviation of each object with respect to its local neighbourhood, is then defined as the the local reachability density of the object $p$ divided by the local reachability density of object $p$'s $MintPts$-nearest neighbour. It is defined as:

\begin{equation}
	LOF_{MinPts}(p)=\frac{{\mathlarger{\sum}_{o\in_{N_{MinPts}(p)}}{\frac{lrd_{MinPts}(o)}{lrd_{MinPts}(p)}}}}{|N_{MinPts}(p)|}
	\label{equ:LOF}
\end{equation}

A low LOF value indicates that the data point is an inlier, while outliers produce greater LOF values, i.e. usually greater than 1.

\subsection{\normalsize \textbf{Prune-based Local Outlier Factor (PLOF) Approach}}
\label{sec:plof}
\par A major drawback of the methods that were proposed previously such as \cite{Poddar2018} and \cite{5734938} is that they all need to cluster the dataset in advance, and then prune a portion of the data instances based on a metric. This kind of approach brings in the complexity of the clustering method into LOF. In our proposed method, there is no need to cluster the dataset.
\par
\begin{figure}
\centering	
\definecolor{xdxdff}{rgb}{0.49019607843137253,0.49019607843137253,1.}
\definecolor{qqffqq}{rgb}{0.,1.,0.}
\definecolor{ududff}{rgb}{0.30196078431372547,0.30196078431372547,1.}
\definecolor{ffqqqq}{rgb}{1.,0.,0.}
\begin{tikzpicture}[line cap=round,line join=round,>=triangle 45,x=0.5333333333333333cm,y=0.5333333333333333cm]
\draw[->,color=black] (-7.5,0.) -- (7.5,0.);
\foreach \x in {-7.,-6.,-5.,-4.,-3.,-2.,-1.,1.,2.,3.,4.,5.,6.,7.}
\draw[shift={(\x,0)},color=black] (0pt,2pt) -- (0pt,-2pt);
\draw[->,color=black] (0.,-7.5) -- (0.,7.5);
\foreach \y in {-7.,-6.,-5.,-4.,-3.,-2.,-1.,1.,2.,3.,4.,5.,6.,7.}
\draw[shift={(0,\y)},color=black] (2pt,0pt) -- (-2pt,0pt);
\clip(-7.5,-7.5) rectangle (7.5,7.5);
\draw [line width=2.pt] (0.,0.) circle (3.2cm);
\draw [line width=2.pt] (0.,0.)-- (5.915555615116677,-1.0030960893461596);
\begin{scriptsize}
\draw [fill=ffqqqq] (0.,0.) circle (4.5pt);
\draw[color=ffqqqq] (-0.4, -0.4) node {$A$};
\draw [fill=ududff] (-3.,1.) circle (2.5pt);
\draw [fill=ududff] (-2.,1.) circle (2.5pt);
\draw [fill=ududff] (-2.,2.) circle (2.5pt);
\draw [fill=ududff] (-4.,3.) circle (2.5pt);
\draw [fill=ududff] (-1.,-1.) circle (2.5pt);
\draw [fill=ududff] (-2.,-2.) circle (2.5pt);
\draw [fill=ududff] (-3.,-3.) circle (2.5pt);
\draw [fill=ududff] (-1.,-3.) circle (2.5pt);
\draw [fill=ududff] (1.,-3.) circle (2.5pt);
\draw [fill=ududff] (2.,-4.) circle (2.5pt);
\draw [fill=ududff] (1.,-1.) circle (2.5pt);
\draw [fill=ududff] (2.,-1.) circle (2.5pt);
\draw [fill=ududff] (2.,-2.) circle (2.5pt);
\draw [fill=qqffqq] (5.915555615116677,-1.0030960893461596) circle (4.5pt);
\draw[color=qqffqq] (6,-1.5) node {$g$};
\draw [fill=xdxdff] (5.0347497149025315,3.2636322262609307) circle (2.5pt);
\draw [fill=ududff] (4.,1.) circle (2.5pt);
\draw [fill=ududff] (1.,3.) circle (2.5pt);
\draw [fill=ududff] (3.,1.) circle (2.5pt);
\draw [fill=ududff] (-1.,3.) circle (2.5pt);
\draw [fill=ududff] (-2.44,3.42) circle (2.5pt);
\draw[color=black] (2.95826071042396,-0.8) node {$r$};
\end{scriptsize}
\end{tikzpicture}

\caption{Neighbourhood of A based on $k$-distance}
\label{fig:loca_density_of_A}
\end{figure}
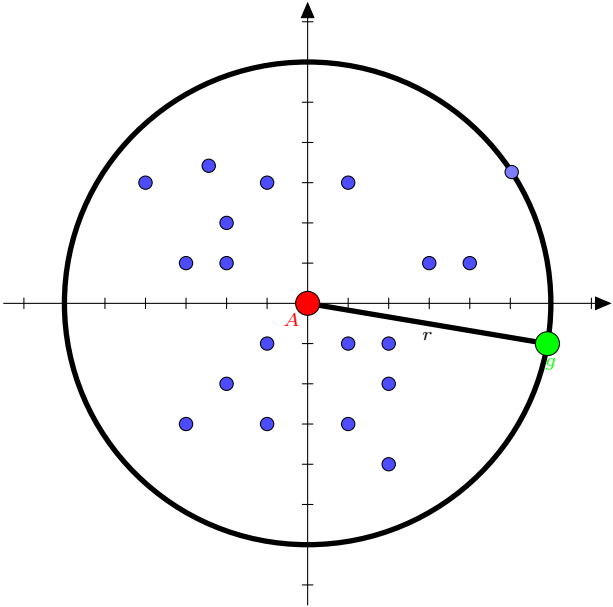

 As explained in Section \ref{alg:lof}, to compute a score by LOF, which indicates the outlierness of a data instance, first need to find out the reachability distance, then local reachability distance, and finally local outlier factor. In our proposed work, these three steps are carried out merely on data instances that requires further investigation. This is decided based on a pruning method. The complexity of this method is $N$, where $N$ is the number of instances. In our pruning method, the local density of each data instance is estimated based on the following equation:
\begin{equation}
	\delta = \frac{{|M|}^2}{\sum\limits_{i=0}^{n}dist(p,i)}
	\label{equ:delta}
\end{equation}
where $|M|$ is the cardinality of $p$'s neighbours and $dist(p,i)$ is the distance between $p$ and its $i$-th neighbour. As depicted in Fig. \ref{fig:loca_density_of_A}, we draw a circle around point $p$ based on $k$-distance (i.e., the radius that within that we can have at least $k$ neighbours), and then we compute $M$. To the best of our knowledge the Equation \ref{equ:delta} has never been proposed before. To remove the effect of extreme values \cite{8291963}, the largest and the smallest values of $\delta$ are eliminated. 

\par Having a set of $\delta$ values, the median is determined and based on the assumption that outliers should have a low $\delta$ value, all the instances that have a $\delta$ value less than the median are pruned. This simple median-based clustering eliminates the problem of finding a good threshold. It is worth to mention that pruned instances are still used for computing Equations \ref{equ:reach-dist}, \ref{equ:lrd}, and \ref{equ:LOF} for not pruned instances. Finally, the LOF value for points that their $\delta$ value is greater than the median is assigned to 0.

\begin{algorithm}
\label{alg:lof}
\SetAlgoLined
 
 \SetKwInOut{Input}{input}\SetKwInOut{Output}{output}
 \Input{$D$: a dataset with $n\times m$ dimension}
 \Output{$\lambda$: a $1d$ array} 
 \BlankLine
 
 initialise: $k = minimum\ neighbours$\;
 $S$ = $\delta$ values\;
 \For{each point $x_i$ in $D$}{
	$NN_i$ = find the $k$-th neighbour $x_j$ for $x_i$\;
	$\delta_i$ = compute $\delta$ for $x_i$ given $NN_i$\;	
	append $\delta_i$ to $S$\;  	
 }
 eliminate extreme values of $\delta$\;
 $med$ = median of $S$\;
 \For{each $x_i$ in $D$}{
 	\If{$\delta_i < med$}{
 	 prune $x_i$
 	}
 }
 \For{each $x_i$ in $D$}{
 	\eIf{$x_i$ not pruned}{
 		compute reachability-distance of $x_i$\;
 		compute local reachability density of $x_i$\;
 		$\rho_i$ = compute LOF value of $x_i$\;
 		append $\rho_i$ to $\lambda$\;
 		}{
 		append $0$ to $\lambda$;
 	}
 }
 \KwRet{$\lambda$}
 \caption{Efficient LOF Algorithm}
\end{algorithm}

A paper should have a short, straightforward title directed at general readers in no more than 20 words.

\section{\Large{Complexity Analysis}}
\label{sec:complexity}
\par The complexity depends on the use of KD-tree. In case of using KD-tree, the time complexity of finding the nearest neighbours is $O(N*\log N)$, where $N$ is the number of instances in the dataset. In case of using a Brute-Force for getting the nearest neighbour, the complexity becomes $O(N^2)$. 
\par The complexity of computing $\delta$ value of each data instance is $O(N)$ while the complexity of getting median of $\delta$ values is $O(1)$. And, the complexity of LOF is $O(N^2)$ \cite{8291963}. By pruning, the size of $N$ reduces, i.e., ($N^{\prime} \ll N$), therefore, the complexity of computing LOF is substantially reduced.

%
%
\begin{table}[h!]
	\caption{The details of 6 datasets}
	\label{table:details}
	\begin{center}
	\footnotesize
		\begin{tabular}{l c c c c }
			\hline
			Dataset & \# of features & \# of outliers & \# of data \\
			\hline
			Wine & 13 & 10 & 129 \\
			Lymphography & 18 & 6 & 148 \\
			Glass & 9 & 9 & 214 \\
			Ionosphere & 33 & 126 & 351 \\
			WBC & 30 & 21 & 278 \\
			Heart & 22 & 15 & 187 \\
			Breast & 9 & 239 & 683 \\
			\hline
		\end{tabular}
	\end{center}
\end{table}

\begin{table}[h!]
	\caption{The execution time}
	\label{table:consuming_time}
	\begin{center}
	\footnotesize
		\begin{tabular}{l c c c c }
			\hline
			Dataset & PLOF & LOF & devToMean & FastLOF \\
			\hline
			Wine & 0.156 & 0.229 & 0.335 & 0.067 \\
			Lymphography & 0.207 & 0.396 & 0.215 & 0.079 \\
			Glass & 0.415 & 0.576 & 0.425 & 0.113 \\
			Ionosphere & 1.167 & 1.297 & 1.081 & 0.214 \\
			WBC & 1.320 & 1.719 & 1.243 & 0.238 \\
			Heart & 0.340 & 0.445 & 0.358 & 0.125 \\
			Breast & 4.336 & 5.163 & 4.335 & 0.633 \\
			\hline
			Average & 1.134 & 1.403 & 1.141 & \textbf{0.209} \\
			\hline
		\end{tabular}
	\end{center}
\end{table}

\begin{table}[h!]
	\caption{The accuracy}
	\label{table:accuracy}
	\begin{center}
		\footnotesize
		\begin{tabular}{l c c c c }
			\hline
			Dataset & PLOF & LOF & devToMean & FastLOF \\
			\hline
			Wine & 0.783 & 0.946 & 0.922 & 0.682 \\
			Lymphography & 0.743 & 0.743 & 0.926 & 0.743 \\
			Glass & 0.734 & 0.659 & 0.921 & 0.715 \\
			Ionosphere & 0.877 & 0.638 & 0.661 & 0.678 \\
			WBC & 0.757 & 0.646 & 0.915 & 0.685 \\
			Heart & 0.572 & 0.519 & 0.578 & 0.540 \\
			Breast & 0.851 & 0.613 & 0.630 & 0.848 \\
			\hline
			Average & 0.759 & 0.680 & \textbf{0.793} & 0.698 \\
			\hline
		\end{tabular}
	\end{center}
\end{table}

\begin{table}[h!]
	\caption{The precision}
	\label{table:precision}
	\begin{center}
		\footnotesize
		\begin{tabular}{l c c c c }
			\hline
			Dataset & PLOF & LOF & devToMean & FastLOF \\
			\hline
			Wine & 0.263 & 0.714 & 0.500 & 0.030 \\
			Lymphography & 0.136 & 0.136 & 0.143 & 0.056 \\
			Glass & 0.125 & 0.000 & 0.100 & 0.036 \\
			Ionosphere & 0.895 & 0.495 & 0.706 & 0.566 \\
			WBC & 0.186 & 0.000 & 0.000 & 0.062 \\
			Heart & 0.536 & 0.446 & 0.778 & 0.481 \\
			Breast & 0.701 & 0.463 & 0.150 & 0.857 \\
			\hline
			Average & \textbf{0.406} & 0.322 & 0.339 & 0.298 \\
			\hline
		\end{tabular}
	\end{center}
\end{table}

\begin{table}[h!]
	\caption{The AUC}
	\label{table:AUC}
	\begin{center}
		\footnotesize
		\begin{tabular}{l c c c c }
			\hline
			Dataset & PLOF & LOF & devToMean & FastLOF \\
			\hline
			Wine & 0.882 & 0.882 & 0.637 & 0.416\\
			Lymphography & 0.866 & 0.866 & 0.562 & 0.547 \\
			Glass & 0.808 & 0.344 & 0.534 & 0.479 \\
			Ionosphere & 0.849 & 0.589 & 0.537 & 0.627 \\
			WBC & 0.871 & 0.342 & 0.485 & 0.520 \\
			Heart & 0.552 & 0.498 & 0.532 & 0.518 \\
			Breast & 0.885 & 0.624 & 0.487 & 0.809 \\
			\hline
			Average & \textbf{0.816} & 0.592 & 0.539 & 0.559 \\
			\hline
		\end{tabular}
	\end{center}
\end{table}

\begin{table}[h!]
	\caption{The recall}
	\label{table:recall}
	\begin{center}
		\footnotesize
		\begin{tabular}{l c c c c }
			\hline
			Dataset & PLOF & LOF & devToMean & FastLOF \\
			\hline
			Wine & 1.0 & 1.0 & 0.300 & 0.100 \\
			Lymphography & 1.0 & 1.0 & 0.167 & 0.333 \\
			Glass & 0.889 & 0.000 & 0.111 & 0.222 \\
			Ionosphere & 0.746 & 0.413 & 0.095 & 0.444 \\
			WBC & 0.905 & 0.000 & 0.000 & 0.333 \\
			Heart & 0.357 & 0.298 & 0.083 & 0.298 \\
			Breast & 1.0 & 0.661 & 0.013 & 0.678 \\
			\hline
			Average & \textbf{0.842} & 0.481 & 0.109 & 0.344 \\
			\hline
		\end{tabular}
	\end{center}
\end{table}

\begin{figure*}[h!]
    \centering
    \begin{subfigure}[t]{0.5\textwidth}
        \centering
        \includegraphics[width=\textwidth]{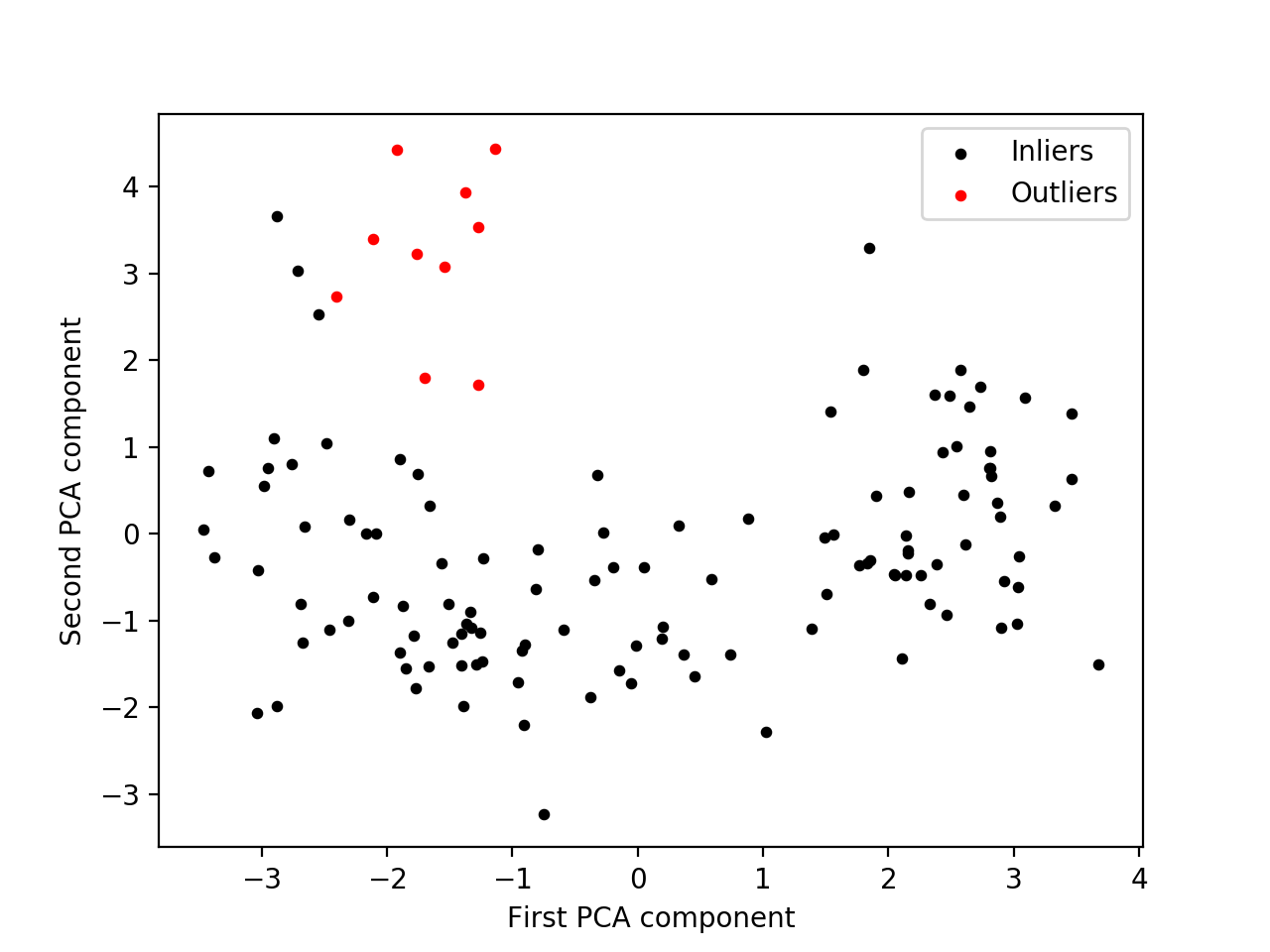}
    \end{subfigure}~
    \begin{subfigure}[t]{0.5\textwidth}
        \centering
        \includegraphics[width=\textwidth]{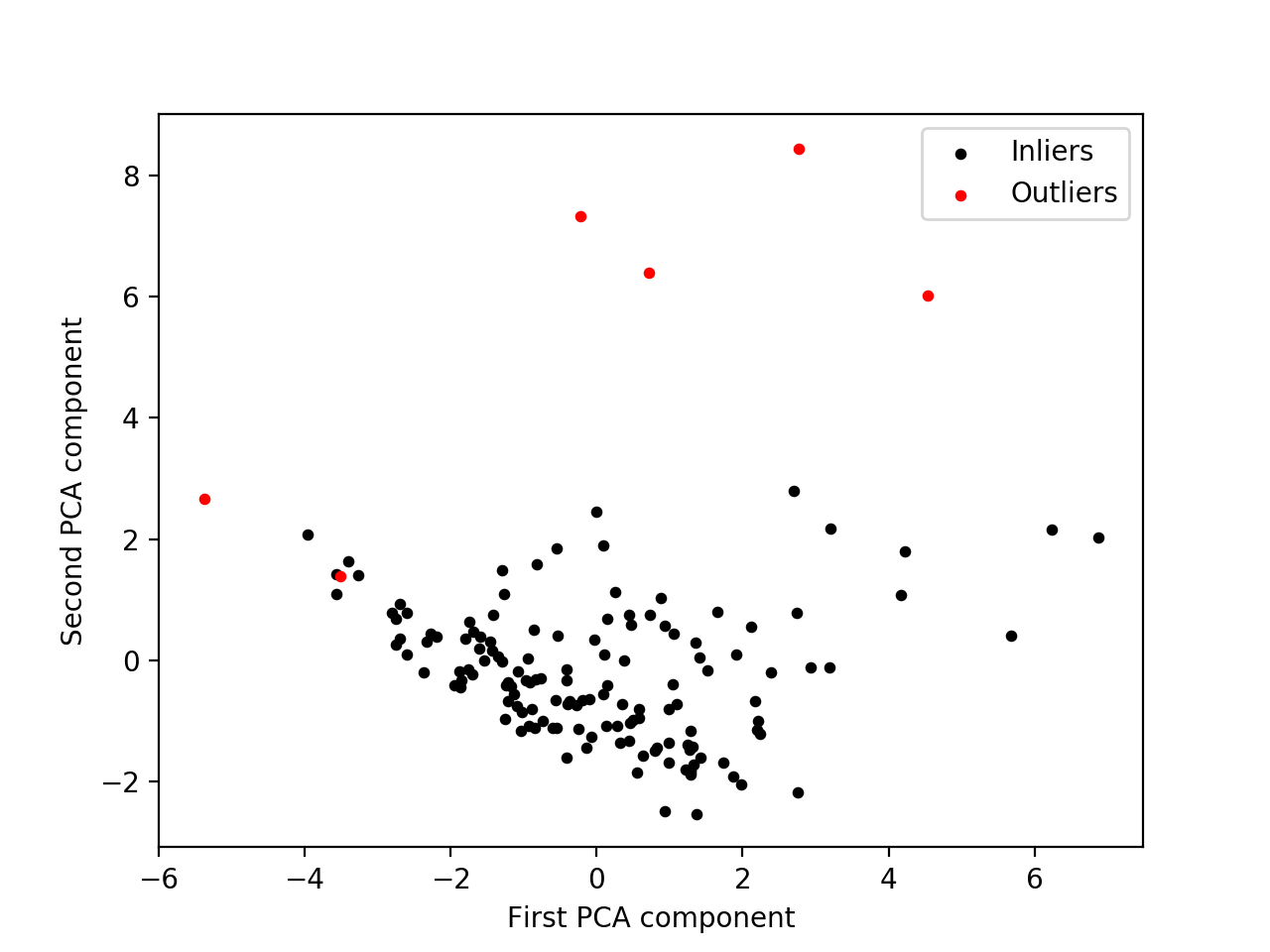}
    \end{subfigure}~
    
    \begin{subfigure}[t]{0.5\textwidth}
        \centering
        \includegraphics[width=\textwidth]{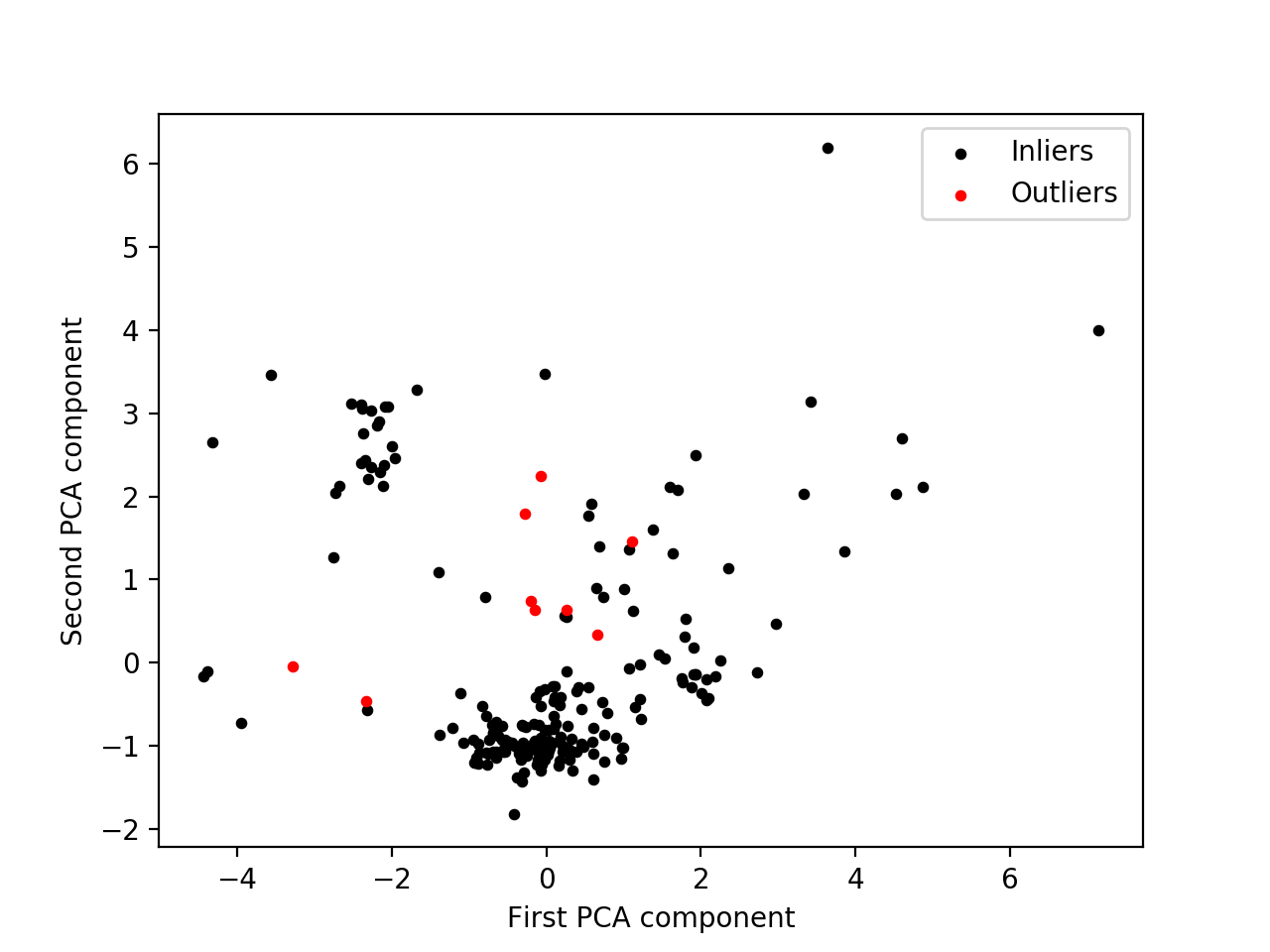}
    \end{subfigure}~
    \begin{subfigure}[t]{0.5\textwidth}
        \centering
        \includegraphics[width=\textwidth]{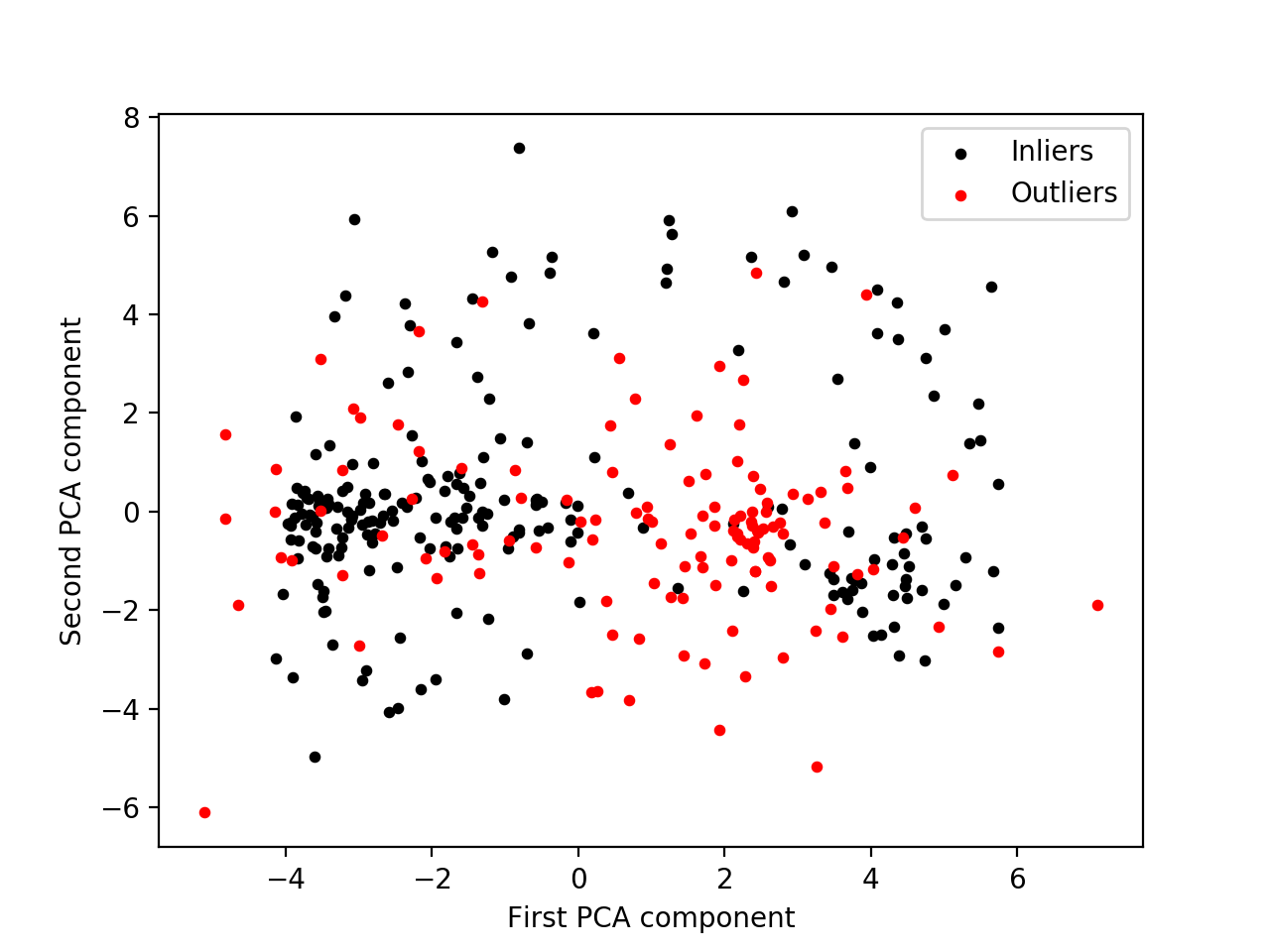}
    \end{subfigure}
    
    \caption{\textbf{Showing the first two principal components of (A) Wine, (B) Lymphography, (C) Glass, and (D) Ionosphere}}
\end{figure*}

\section{\Large{Experiment Results and Analysis}}
\label{sec:experiment}
\par The proposed method is conducted on seven real datasets obtained from UCI Machine Learning Repository \cite{Dua:2017}, which are all publicly available. In the context of outlier detection, the datasets that are chosen in this paper are considered as the benchmark as they have been frequently used in the literature. The details of each dataset that is used in our experiment are given in Table \ref{table:details}. It is worth to mention that because most of these datasets are originally used in classification, some of the datasets are modified, e.g., merging one or two major classes to form the normal class. 

\par Various evaluation metrics have been used in the literature, however, the most effective and widely used metric, especially for evaluating unsupervised methods, is Receiver Operator Characteristics (ROC) curve, which basically shows the true positive rate versus the false negative rate at different threshold \cite{Tang2017171}. By calculating the Area Under ROC (AUC) curve, it is possible to show the effectiveness of the algorithm by a single value. The value of AUC can vary between 0 to 1 in which any value closer to 1 indicates better performance while 0.5 shows a random decision making. In this paper, we have used AUC, and also other evaluation metrics such as accuracy, precision, and recall \cite{Campos20151}. The experiment on each dataset is carried out five times and the averaged outcome is reported. 

\par We compared our approach with the original LOF algorithm \cite{Breunig:2000:LID:335191.335388}, and also another two similar approaches that were published recently: FastLOF \cite{6460620} and devToMean \cite{Poddar2018}. The two latter approaches are selected because their main purpose is also to reduce the execution time of LOF. The original LOF requires merely one parameter which is the minimum number of neighbours. But, it is possible to also define a threshold and pick data instances that their LOF value exceeds the threshold. In our experiment, we have used the same parameters for each approach.

\par In terms of execution time and accuracy, our approach came second after FastLOF and devToMean respectively. However, by looking at other metrics one can deduce that \cite{6460620} and \cite{Poddar2018} sacrificed detection performance, which is the main goal, for execution time. For instance, our approach is the dominating method based on AUC with a relatively big margin and a standard deviation of $0.119$. Moreover, PLOF proved to be superior, on average, based on recall and precision achieving $0.842$ and $0.406$ respectively. 

\par By using PLOF, not only we managed to reduce execution time but also improved detection performance. 

\section{\Large{Conclusion}}
\label{sec:conclusion}
\par Prune-based local outlier factor reduces the execution time of LOF. By introducing a novel pre-processing step, data instances with high density are pruned and LOF is only applied to unpruned data instances. In this way, the complexity of LOF reduces while its performance remains unchanged. Based on our experiment that was conducted on 8 real-life datasets, PLOF demonstrated superior results compared with the original LOF and two more state-of-the-art variants of LOF. In the future, we plan to apply our novel method to other density-based methods to evaluate its performance.

\noindent\hrulefill

\renewcommand\refname{REFERENCES}

\bibliography{main.bib}{}
\bibliographystyle{ieeetr}

\end{document}